\documentclass[letterpaper, 10 pt, conference]{ieeeconf}
\IEEEoverridecommandlockouts
\usepackage{bobbystyle_IEEEjournal}

\usepackage{xcolor}
\usepackage{graphicx}
\usepackage{amsfonts}
\usepackage{amssymb}
\usepackage{amsmath}
\usepackage{booktabs}   
\usepackage{multirow}   
\usepackage{array}      
\usepackage{color}
\usepackage{colortbl}
\usepackage{makecell}
\usepackage{cite}
\usepackage{multirow}
\usepackage{ulem}
\makeatletter
\let\NAT@parse\undefined
\makeatother
\usepackage{hyperref}
\hypersetup{
    colorlinks=true,
    linkcolor=blue,
    filecolor=blue,      
    urlcolor=blue,
    citecolor=blue
}

\newcommand{\dquotes}[1]{``#1''}

\usepackage{tikz}
\newcommand{\mycirc}[1]{\textcircled{\scriptsize #1}}

\NewDocumentCommand{\boldbigcirc}{O{1.15ex} O{0.6pt}}{%
  \mathbin{\tikz[baseline=-0.6ex]{\draw[line width=#2] (0,0) circle (#1);}}%
}

\title{\LARGE \bf 
SToRM: Supervised Token Reduction for Multi-modal LLMs \\ 
toward efficient end-to-end autonomous driving}

\author{Seo Hyun Kim, Jin Bok Park, Do Yeon Koo, Hogun Park$^{\dagger}$, Il Yong Chun$^{\dagger}$%
\thanks{
$^\dag$Corresponding authors.
The work was supported in part by NRF Grant RS-2023-00213455 funded by MSIT, 
the Digital Therapeutics Development and Demonstration Support Program funded by MSIT and NIPA, 
the BK21 FOUR Project, 
IITP Grant RS-2019-II190421 (AI Graduate School Support Program (Sungkyunkwan University)) funded by MSIT, 
KIAT Grant RS-2024-00418086 (HRD Program for Industrial Innovation) funded by MOTIE, 
and IBS-R015-D1.
}
\thanks{Seo Hyun Kim, Jin Bok Park and Do Yeon Koo are with the Department of Electrical and Computer Engineering (ECE), Sungkyunkwan University (SKKU), Suwon 16419, South Korea (email: rlatjgus0608@g.skku.edu, bjb663@g.skku.edu, kdy1021@g.skku.edu).}
\thanks{Hogun Park is with the Department of Computer Science Engineering, Artificial Intelligence (AI), and Intelligent Software, SKKU, Suwon 16419, South Korea (email: hogunpark@skku.edu).}
\thanks{Il Yong Chun is with the Departments of ECE, AI, Advanced Display Engineering, and Semiconductor Convergence Engineering, Display Convergence Engineering, SKKU, Suwon 16419, South Korea, and also with the Center for Neuroscience Imaging Research, Institute for Basic Science, Suwon 16419, South Korea (email: iychun@skku.edu).}
}

\begin{document}
\maketitle
\thispagestyle{empty}
\pagestyle{empty}

\begin{abstract}
In autonomous driving, end-to-end (E2E) driving systems that predict control commands directly from sensor data achieved significant advancements.
For safe autonomous driving in unexpected scenarios,
one may additionally rely on human interventions such as natural language instructions.
Using a multi-modal large language model (MLLM) in autonomous driving 
facilitates human–vehicle interactions, and may improve driving performances in unexpected scenarios.
However, this approach requires substantial computational resources due to its reliance on an LLM and many visual tokens from sensor inputs, that are inherently limited in autonomous vehicles.
Many MLLM studies have explored reducing the number of visual tokens, and many approaches tend to exhibit some end-task performance degradation compared to using all tokens.
For efficient E2E driving while maintaining driving performance comparable to using all tokens, 
this paper proposes the first \emph{S}upervised \emph{To}ken \emph{R}eduction framework for \emph{M}ulti-modal LLMs (\emph{SToRM}).
The proposed SToRM framework consists of three key elements.
First, we propose a lightweight \emph{importance predictor} with short-term sliding windows that predicts the importance scores of visual tokens. 
Second, we propose a supervised learning approach for the importance predictor, that uses an auxiliary path to obtain \emph{pseudo-supervision signals} from an all-token pass through the LLM.
Third, guided by predicted importance scores, we propose an \emph{anchor–context merging} module that partitions tokens into \dquotes{anchors} and \dquotes{context} tokens, then merges the latter into their most relevant anchors to reduce redundancy while minimizing information loss.
Experiments with the LangAuto benchmark dataset show that the proposed SToRM outperforms state-of-the-art E2E driving MLLM under an equal reduced-token budget and maintains all-token performance while substantially reducing computational cost, by up to $30\times$, and enabling real-time E2E driving on a standard GPU.
\end{abstract}

\section{Introduction}
\label{sec:intro}

The end-to-end (E2E) driving approach that directly transforms sensor data to control signals has achieved significant advancements in autonomous driving \cite{e2e_survey}.
However, E2E driving methods often face challenges in complex and unforeseen scenarios that need high-level reasoning and contextual understanding. 
For safe autonomous driving, human intervention through natural language instructions is critical, by facilitating interactions between human decisions and vehicles \cite{llm_ad_survey}.
For example, in scenarios where a pedestrian suddenly appears, an autonomous vehicle can promptly adapt to such scenario by additionally understanding the language instructions from a driver.
By leveraging language instructions, one may mitigate the inherent generalization limitation in E2E driving, ultimately achieving safer driving in unforeseen scenarios \cite{LMDrive, Generalization_E2E, drivelm}.

\begin{figure}[t!]
    \centering
    \includegraphics[width=\linewidth]{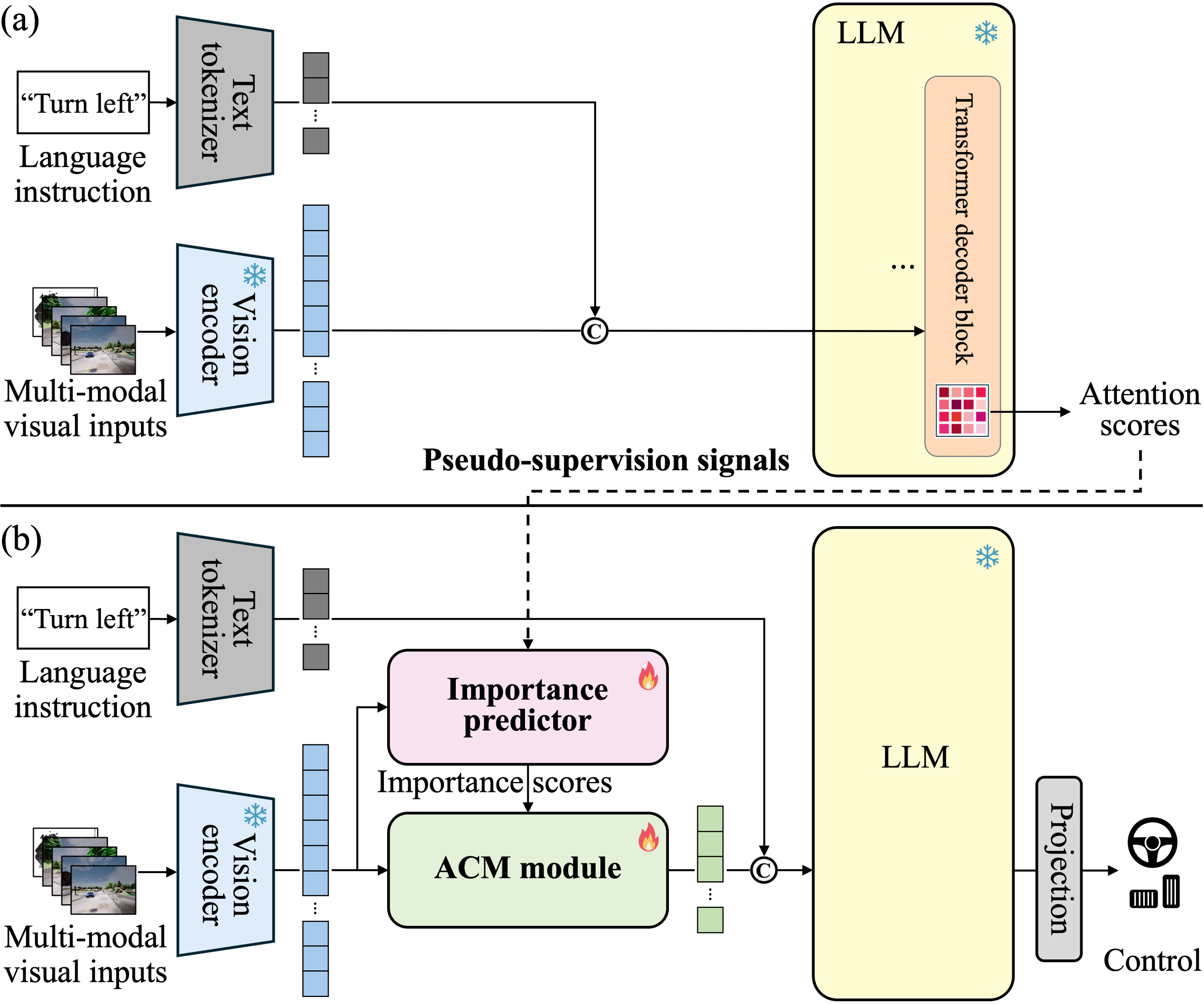}

    \vspace{-0.5pc}
    \caption{Overview of the proposed SToRM framework.
    (a) SToRM is built on the central idea of leveraging intermediate results from an MLLM -- specifically, attention scores derived from all tokens -- as {\bfseries pseudo-supervision signals} for training an importance predictor to reduce visual tokens.
    (b) In addition, we propose \textit{\romnum{1})} a new lightweight {\bfseries importance predictor} learned from pseudo-supervision signals; and \textit{\romnum{2})} an {\bfseries anchor-context token merging (ACM)} module that reduces visual tokens while preserving essential information. 
    We train SToRM in an E2E manner.}
    \label{fig:overview}
\vspace{-1pc}
\end{figure}

With the success of large language models (LLMs) in language understanding, multi-modal LLMs (MLLMs) have been proposed to compensate for modality-specific limitations and to harness complementary information across modalities.
MLLMs integrate information from diverse inputs (e.g., text, images, audio) by projecting modality-specific features into a shared representation space and processing aggregated tokens with an LLM backbone.
This architecture enables joint reasoning across modalities and facilitates alignment between visual and textual information.
In autonomous driving, MLLMs can leverage complementary information from external sensors together with language inputs such as navigation instructions.

However, applying MLLMs to E2E autonomous driving systems has significant limitations.
In E2E autonomous driving, temporal reasoning over historical information can improve driving performance by fusing features across multiple frames \cite{ReasonNet}.
Yet, processing several past frames through a vision encoder produces a substantially larger number of visual tokens than text tokens.
Leveraging many visual tokens may improve driving performance, but it incurs a considerable computational burden,
since LLMs consist of numerous attention layers whose computational complexity grows quadratically with input length \cite{Transformer}.
This severely degrades inference speed, which is critical for autonomous vehicles that demand real-time operation.
To mitigate this issue, the prior work \cite{LMDrive} reduced the number of visual tokens by Q-Former, a query-based transformer module \cite{BLIP2}.
Although this approach lowers computational cost, we observed that it often results in degraded E2E driving performance compared to using all visual tokens.

To reduce computational costs in E2E driving while maintaining performance comparable to using all tokens, this paper proposes the first \textbf{S}upervised \textbf{To}ken \textbf{R}eduction framework for \textbf{M}ulti-modal LLMs (\textbf{SToRM}).
The central idea behind SToRM is to leverage intermediate results from an MLLM, particularly attention scores derived from all tokens, as pseudo-supervision signals for training an importance predictor to reduce visual tokens.
This proposed mechanism is based on the assumption that tokens receiving higher attention in an LLM indicate greater importance.
The proposed SToRM framework consists of three key elements: 
\textit{\romnum{1})} constructing pseudo-supervision signals (see above); 
\textit{\romnum{2})} importance predictor for all visual tokens, and 
\textit{\romnum{3})} anchor-context token merging (ACM) module.
See the overview of SToRM in Fig.~\ref{fig:overview}.
To predict importance scores of visual tokens with low computational costs, 
we propose a new lightweight importance predictor that captures short-term spatio-temporal relations among visual tokens and intra-token cross-channel dependencies, designed via a new architecture of Multi-Layer Perceptron-Mixer (MLP-Mixer) \cite{mlp_mixer} block with a temporal sliding window.
To reduce visual tokens while minimizing information loss, 
we propose an ACM module that partitions visual tokens into \dquotes{anchor} and \dquotes{context} groups based on predicted importance, and then merges each context token into its most relevant anchor.

Our main contributions can be summarized as follows:
\begin{itemize}
    \item We propose \textbf{SToRM}, the \emph{first} supervised token-reduction framework for MLLMs in E2E driving, that leverages pseudo-supervision signals to guide importance-aware token reduction.
    
    \item We propose a lightweight importance predictor that captures \emph{short-term} spatio-temporal relations among visual tokens rather than long-range dependencies spanning the entire token sequence; it also models intra-token cross-channel dependencies.
    
    \item We propose an ACM module that reduces the number of visual tokens by merging each context token into its most relevant anchor, with anchors selected based on high predicted importance scores.

    \item Our experiments with the LangAuto benchmark \cite{LMDrive} show that the proposed SToRM outperforms state-of-the-art (SOTA) E2E driving MLLM under an equal token budget and maintains all-token performance, while substantially reducing computational resources.
\end{itemize}

\vspace{-0.1pc}
\section{Related works}
\label{sec:related works}

To handle challenging scenarios in E2E autonomous driving, one may use an MLLM with both visual and language information, promoting interactions between human decisions and information from multiple sensors.
In general, such MLLMs suffer from a large number of visual tokens generated from raw sensor inputs that leads to quadratic growth in computation cost due to the numerous attention layers in an LLM backbone.
To accelerate the inference speed of E2E driving MLLM, it is crucial to reduce the number of visual tokens.

Unlike Vision Transformers \cite{ViT}, where the number of visual tokens is fixed by patch size and image resolution, MLLMs feed multi-modal tokens into an LLM backbone that is inherently capable of processing variable-length sequences.
Several studies have leveraged this property to reduce visual tokens in MLLMs.

For example, LMDrive, the first E2E driving MLLM, used Q-Former to reduce the number of visual tokens from several sensors \cite{LMDrive}. 
Q-Former uses the cross-attention mechanism between learnable queries and visual tokens, and only these queries are passed to an LLM \cite{BLIP2}.
Beyond autonomous driving, many studies have also explored reducing visual tokens in MLLMs.
For example, HICom is a hybrid-level instruction-guided token compression method that injects instructions into both local and global visual tokens using learnable queries \cite{HICom}.
HiRED is a plug-and-play token-dropping method that select the most informative tokens based on a special classification token (\textsf{[CLS]}) from vision encoder \cite{HiRED}.
Token Merging (ToMe) gradually reduces the number of visual tokens in each transformer block by selecting the most representative tokens via bipartite matching \cite{ToMe}.  
LLaVA-PruMerge adaptively reduces visual tokens by pruning and merging based on similarities between the most representative token and others \cite{PruMerge}.  
VisionZip selects informative visual tokens based on its attention score and merges remaining tokens based on similarity \cite{VisionZip}.
DivPrune is a token pruning method by selecting a subset that the diversity among the selected visual tokens is maximized \cite{DivPrune}.

However, the aforementioned methods reduce visual tokens \emph{without} task supervision, relying on heuristic signals like similarity that led to a trade-off between inference efficiency and task performance.
In contrast, the proposed supervised token reduction framework reduces computational cost \emph{without} sacrificing performance, by supervising token importance with pseudo-importance scores.

\begin{figure*}[t!]
    \centering
    \includegraphics[width=0.9\linewidth]{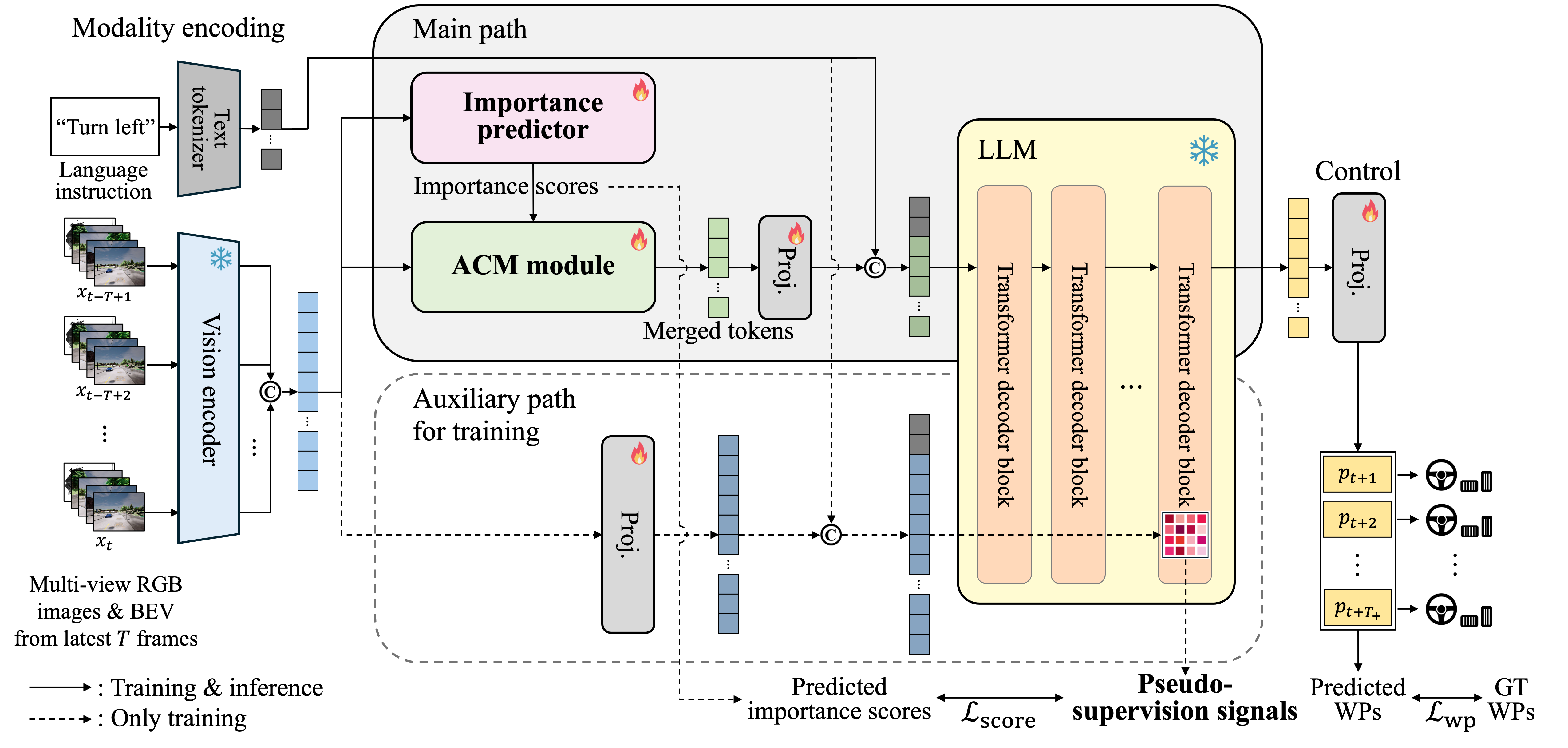}
    \vspace{-0.5pc}
    \caption{The overall architecture of the proposed SToRM framework (at each current time step $t$). The symbol \mycirc{c}, WPs, and GT denote a concatenation operator, waypoints over the next $T_+$ frames, and ground truth, respectively. The overall architecture is described at the beginning of \S\ref{sec:methods}.}
    \label{fig:overall_architecture}
    \vspace{-1pc}
\end{figure*}

\section{Methods}
\label{sec:methods}

The proposed SToRM framework takes red-green-blue (RGB) images from multi-view cameras, a bird's-eye-view (BEV) map derived from point clouds obtained by a light detection and ranging (LiDAR) sensor and language instructions as input, and predicts control commands (e.g., steer, throttle, and brake) as output through an MLLM, while reducing visual tokens.
To capture temporal relations, we use a set of multi-modal visual inputs from $T$ consecutive frames, paired with a language instruction.
Fig.~\ref{fig:overall_architecture} illustrates the overall architecture of SToRM for E2E driving:
\begin{itemize}
    \item {\bfseries Modality encoding:} We employ a frozen vision encoder backbone to extract visual tokens from multi-view RGB images and a BEV map at each frame, and a text tokenizer to extract text tokens from a language instruction.
    We concatenate the visual tokens from $T$ frames to capture temporal driving context.
    
    \item {\bfseries Main path:} The main path consists of two core modules.
    First, we predict importance scores of all visual tokens from a vision encoder by the proposed lightweight {\bfseries importance predictor}.
    We then reduce the number of visual tokens by the proposed {\bfseries ACM module} using predicted importance scores.
    We project the merged tokens, concatenate them with text tokens, and feed them into a frozen LLM to generate output tokens for estimating control commands.
    
    \item {\bfseries Auxiliary path for training:}
    The purpose of this path is to generate {\bfseries pseudo-supervision signals} for training the importance predictor, by using \emph{all} visual tokens \emph{without} token reduction.
    We feed all visual and text tokens into the frozen LLM without token reduction.
    We then use its last decoder's attention scores as pseudo-supervision signals to train the importance predictor.

    \item {\bfseries Control:} At each time step, we predict multiple future waypoints from the output tokens of the frozen LLM using reduced visual tokens, and convert them into control commands.

    \item {\bfseries End-to-end training:} We train the entire SToRM in an E2E way by using two losses: \textit{\romnum{1})} $\mathcal{L}_{\text{score}}$ for training the importance predictor via the auxiliary path for training;
    and \textit{\romnum{2})} $\mathcal{L}_{\text{wp}}$ for training the entire model using predicted waypoints.
\end{itemize}

\subsection{Modality encoding}
\label{sec:modality_encoding}

At each time step $t \in \bbN_{\geq T}$, we generate a set of visual tokens $\{ \mb{Z}_{t'} \in \bbR^{N \times D} : t'  = t - T + 1,\ldots, t \}$ from multi-view RGB images and a BEV map over the most recent $T$ frames, using the vision encoder pretrained by \cite{LMDrive}, where {$N$ and $D$} denote the number of visual tokens for each frame and the embedding dimension of each visual token, respectively.
The vision encoder backbone \cite{LMDrive} consists of two convolutional neural network (CNN) encoders and a fusion transformer.
A single CNN encoder is shared across all views to extract features from RGB images, while the other CNN encoder processes the BEV map.
A fusion transformer then generates visual tokens by fusing the features extracted from the two encoders.
We transform a language instruction into text tokens using the SentencePiece tokenizer.

\subsection{Main path}
\label{sec:main_path}

The main path consists of the following three core modules: 
\textit{\romnum{1})} proposed importance predictor, 
\textit{\romnum{2})} proposed ACM module, and 
\textit{\romnum{3})} an LLM backbone.

\begin{figure*}[t!]
    \centering
    \includegraphics[width=\linewidth]{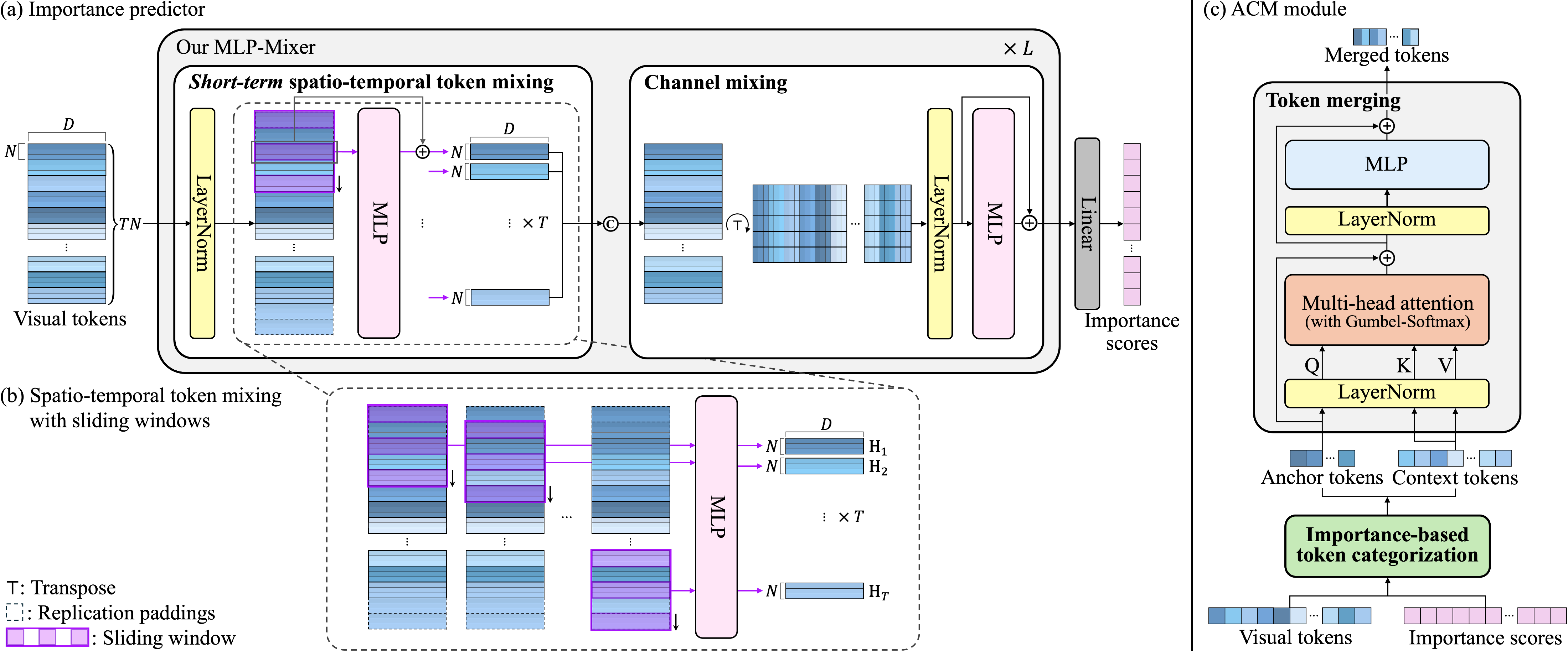}
    \vspace{-1.3pc}
    \caption{The overall architectures of the proposed lightweight importance predictor and ACM module.
    (a) The proposed importance predictor consists of 
    \textit{\romnum{1})} \emph{short-term} spatio-temporal visual token mixing, 
    \textit{\romnum{2})} channel mixing, and 
    \textit{\romnum{3})} importance score computation.
    (b) The proposed \emph{short-term} spatio-temporal visual token mixing mechanism with sliding windows. 
    The input is the entire visual token matrix $\widetilde{\mb{Z}}$ in \R{eq:token_concat}, 
    where the $\tau$th block of $N$ rows indicates the token-embedding matrix $\widetilde{\mb{Z}}_\tau$ at time step $\tau$.
    A purple shaded block denotes a set of short-term spatio-temporal visual tokens selected by a sliding window,
    $\widetilde{\mb{Z}}_{\cW(\tau)}$ in (\ref{eq:window_tokens});
    the one-dimensional checkerboard indicates a dilated sliding window.
    The output of MLP, $\mb{H}_{\tau}$ in (\ref{eq:token_mixing}), 
    represents both spatial structure and short-time temporal evolution at $\tau$.
    (c) The ACM module comprises \textit{\romnum{1})} importance-based token categorization and \textit{\romnum{2})} token merging: we first categorize visual tokens by predicted importance scores from (a), then merge \dquotes{context} tokens into their most relevant \dquotes{anchors} via cross-attention.
    } 
    \label{fig:proposed_modules}
    \vspace{-0.8pc}
\end{figure*}

\subsubsection{Proposed importance predictor}
\label{sec:importance_predictor}

In this section, we propose a lightweight importance predictor that estimates the importance scores of visual tokens from a vision encoder backbone with low computational costs.
Built upon the MLP-Mixer architecture \cite{mlp_mixer}, we propose a new mixing mechanism using sliding windows that captures spatial relations with local temporal context among visual tokens; and we also capture intra-token, cross-channel dependencies.
See the overall architecture of the proposed module in Fig.~\ref{fig:proposed_modules}(a).

At each time step, we aggregate visual tokens across the most recent $T$ frames by concatenating them in a frame-wise manner, followed by layer normalization:
\ea{
\widetilde{\mb{Z}}
&= 
[ \widetilde{\mb{Z}}_1^\top,
\cdots,
\widetilde{\mb{Z}}_T^\top ]^\top
\nn 
\\
&=
\operatorname{LayerNorm}\! 
\left( [ \mb{Z}_{t - T + 1}^\top, 
\cdots, 
\mb{Z}_t^\top ]^\top \right)
\in 
\bbR^{TN \times D}
\label{eq:token_concat}
}
where we omit the time index $t$ in $\widetilde{\mb{Z}}$ for simplicity.
Here, we apply layer normalization, $\operatorname{LayerNorm}(\cdot)$, across the embedding dimension, normalizing each token’s embedding vector to have zero mean and unit variance.
If the token mixing module of the conventional MLP-Mixer \cite{mlp_mixer} is naïvely applied to \R{eq:token_concat}, 
it incurs substantial computational and memory overhead particularly due to the large $TN$
(e.g., $T = 30$ and $N = 100$).
To overcome this limitation, we propose a new spatio-temporal visual token mixing mechanism with sliding windows.

{\bfseries \emph{Short-term} spatio-temporal visual token mixing.}
Rather than considering all visual tokens in $\widetilde{\mb{Z}}$ of \R{eq:token_concat},  
we instead focus on a subset defined within a sliding window: 
\ea{
\widetilde{\mb{Z}}_{\cW(\tau)}
&= 
[ \widetilde{\mb{Z}}_{\tau - \ell \cdot \kappa}^\top, 
\cdots,  
\widetilde{\mb{Z}}_{\tau + \ell \cdot \kappa}^\top
]^\top
\! 
\in \bbR^{| \cW(\tau) | N \times D}
, 
\label{eq:window_tokens}
\\
\cW(\tau) 
&= 
( \tau - \ell \cdot \kappa, \; \dots,\; \tau, \;\dots,\; \tau + \ell \cdot \kappa ),
\nn
}
for $\tau = 1,\ldots,T$,
$\ell \in \bbN_{>0}$ denotes the window radius, such that the window size is given by $2\ell + 1$ (e.g., $\ell=1$ for a window size of $3$ and $\ell=2$ for a window size of $5$), and $\kappa \in \bbN_{[1,\ell+1]}$ denotes the dilation factor.
Since the window may exceed the valid range at sequence boundaries, 
we adopt \emph{replication padding}: out-of-bound indices are set equal to the nearest valid frame, i.e.,
$\widetilde{\mb{Z}}_k = \widetilde{\mb{Z}}_{1}$ if $k < 1$ 
and 
$\widetilde{\mb{Z}}_k = \widetilde{\mb{Z}}_{T}$ if $k > T$. 
This ensures that the window size remains fixed at $2\ell+1$.
We remark that $| \cW(\tau) | = | \cW |$, $\forall \tau$, i.e., the input dimension of a subsequent MLP is identical for $\{ \widetilde{\mb{Z}}_{\cW(\tau)} : \forall \tau \}$, where $| \cdot |$ denotes the length of a sequence.

Now, we design an MLP to mix visual tokens within each sliding window, effectively capturing \emph{short-term} spatio-temporal dependencies:
\be{
\mb{H}_{\tau} 
= 
\widetilde{\mb{Z}}_{\tau}
+ \mb{W}_2 \cdot
\sigma 
\big(\mb{W}_1 \widetilde{\mb{Z}}_{\cW(\tau)} \big)
\in \bbR^{N \times D},
\label{eq:token_mixing}
}
for $\tau = 1,\ldots,T$,
where $\widetilde{\mb{Z}}_{\tau}$ is given in \R{eq:token_concat},
$\sigma(\cdot)$ denotes the GeLU activation,
$\mb{W}_2 \in \bbR^{N \times 2 |\cW| \cdot N}$,
$\mb{W}_1 \in \bbR^{2 |\cW| \cdot N \times |\cW| \cdot N}$,
and $\widetilde{\mb{Z}}_{\cW(\tau)}$ is given in \R{eq:window_tokens}.
Note that $\{ \mb{W}_1, \mb{W}_2 \}$ is shared for all $\tau$.
See an illustration of the proposed processes \R{eq:window_tokens}--\R{eq:token_mixing} in Fig.~\ref{fig:proposed_modules}(b).

Here, $\widetilde{\mb{Z}}_{\mathcal{W}(\tau)}$ concatenates visual tokens from multiple frames 
within the sliding window, 
so the MLP captures both spatial relations (within each frame) and 
temporally local relations (across frames), enabling $\mb{H}_{\tau}$ to encode short-term spatio-temporal context.
For example, the term $\mb{W}_1 \widetilde{\mb{Z}}_{\mathcal{W}(\tau)}$ corresponds to a linear projection of all visual tokens from different spatial locations and time points within a sliding window (to a higher-dimensional space), where each token is modulated by a learnable vector in $\mb{W}_1$.
In effect, the $TN$ spatio-temporal tokens $\{ \mb{H}_{\tau} : \tau \in [1,T] \}$ in \R{eq:token_mixing} are visual tokens that represent both spatial structure and short-time temporal evolution.

\begin{table}[t!]
\centering
\caption{Computational complexity comparison between naïve appl.~of existing token mixing and proposed token mixing}
\label{tab:window_big_o}
\vspace{-0.5pc}
\resizebox{1\columnwidth}{!}{%
\begin{tabular}{c|c}
\toprule
\textbf{Methods} & \textbf{Computation complexity} \\
\midrule
Existing token mixing w/~\R{eq:token_concat} & $\mathcal{O}(D\cdot(TN)^2)$ \\
Proposed token mixing w/~\R{eq:window_tokens} & $\mathcal{O}\!\left(D \cdot (2\ell+1)^2 \cdot (T/\kappa) \cdot N^2 \right)$ \\
\bottomrule
\end{tabular}}
\vspace{-1pc}
\end{table}

The proposed model in \R{eq:token_mixing} is lightweight, as it performs spatio-temporal mixing only within a sliding window rather than across all tokens. 
By restricting interactions to a local temporal context, the model significantly reduces computational and memory overhead while still capturing essential spatio-temporal relations. 
This design choice enables efficient processing of long sequences, since the complexity grows with the window size rather than the entire sequence length.
See Table~\ref{tab:window_big_o} for computational complexity comparison between a naïve application of existing token mixing \cite{mlp_mixer} using \R{eq:window_tokens} and the proposed token mixing mechanism using slide windows via \R{eq:token_mixing}.
Consequently, the sliding-window token mixing offers a favorable trade-off between efficiency and performance, 
making it a practical choice for latency-sensitive autonomous driving settings.
We will later demonstrate that our approach \R{eq:token_mixing} offers computational advantages and favorable driving performance trade-offs over an naïve application of existing token mixing in \cite{mlp_mixer}.

{\bfseries Mixing across the embedding dimension, a.k.a., channel mixing.}
Next, we introduce channel mixing to model how individual feature dimensions (channels) interact across temporally and spatially connected token sequences,
complementing the spatio-temporal token mixing mechanism \R{eq:token_mixing}.
This operation can enrich the representation of each token in $\{ \mb{H}_{\tau} : \forall \tau \}$ in \R{eq:token_mixing}, by capturing dependencies that span along the embedding dimension, which are not addressed by token-level mixing \R{eq:token_mixing} alone.

We aggregate transposed spatio-temporal token representations from $T$ recent frames by concatenating them along the token dimension, followed by layer normalization in \R{eq:token_concat}:
\ea{
\widetilde{\mb{H}}
&= 
\operatorname{LayerNorm}\! 
\left( [\mb{H}_1^\top,
\cdots,
\mb{H}_T^\top] \right)
\in 
\bbR^{D \times TN},
\label{eq:channel_mixing_concat}
}
with $\{ \mb{H}_{\tau} : \tau \in [1,T] \}$ from \R{eq:token_mixing}.
Each row of the constructed matrix $\widetilde{\mb{H}}$ by \R{eq:channel_mixing_concat} corresponds to a 
single embedding channel and spans the entire token sequence (all spatial locations across all $T$ frames). 
For a given channel $d \in \bbN_{[1,D]}$, the row vector 
$\tilde{\mb{h}}_{d}^\top \in \mathbb{R}^{TN}$ 
traces how that feature dimension responds as the viewpoint changes and time progresses. 
Channel mixing learns a function over this per-channel sequence to capture dependencies along the token axis.

We model an MLP for channel mixing, similar to \cite{mlp_mixer}:
\be{
\mb{U} 
= 
\widetilde{\mb{H}}
+ \mb{W}_4 \cdot
\sigma 
\big(\mb{W}_3 \widetilde{\mb{H}} \big)
\in \bbR^{D \times TN},
\label{eq:channel_mixing}
}
where $\widetilde{\mb{H}}$ is given in (\ref{eq:channel_mixing_concat}), 
$\mb{W}_4 \in \bbR^{D \times 2D}$, and
$\mb{W}_3 \in \bbR^{2D \times D}$.
Concretely, we apply the two-layer MLP with a residual connection in \R{eq:channel_mixing} independently to each token vector (i.e., each column of $\widetilde{\mb{H}}$ in \R{eq:channel_mixing_concat}), thereby mixing embedding channels within the token.
This operation captures intra-token, cross-channel dependencies, whereas inter-token (sequence-wise) dependencies are captured by the preceding spatio-temporal token mixing module \R{eq:token_mixing}.
The representation $\mb{U}$ in \R{eq:channel_mixing} thus can encode not only spatial-temporal information but also cross-channel dependencies, enriching the original visual token representations $\{ \mb{Z}_{\tau} : \forall \tau \}$ in \R{eq:token_concat}.
We repeat the sequence of \R{eq:token_concat}–\R{eq:channel_mixing} for $L$ blocks.

We will later show, in the context of predicting importance scores, that the proposed sliding window–based MLP mixer blocks \R{eq:token_mixing} and \R{eq:channel_mixing} effectively avoid the quadratic computational complexity of conventional Transformer blocks.

{\bfseries Importance score computation.}
We finally compute the importance score of each token that quantifies its contribution to downstream decisions, by applying a linear projection:
\be{
\mb{s}^\top
= 
\mb{W}_5
\mb{U}
\in \bbR^{1 \times TN},
\label{eq:final_score}
}
where $\mb{U}$ is given in (\ref{eq:channel_mixing}) and $\mb{W}_5 \in \bbR^{1 \times D}$.

\subsubsection{Proposed ACM module}
\label{sec:ACM}

In this section, we propose an ACM module that, 
at each time step $t' \in [t - T + 1, t]$, merges \dquotes{context} tokens into a group of \dquotes{anchor} tokens that are selected based on high predicted importance scores.
Given visual tokens $\widetilde{\mb{Z}}$ in (\ref{eq:token_concat}) and the corresponding importance $\mb{s}$ in (\ref{eq:final_score}) over recent $T$ frames, 
proposed ACM operates independently at each frame by defining the followings:
$\widetilde{\mb{Z}}_{\tau} \in \bbR^{N \times D}$ and $\mb{s}_{\tau} \in \bbR^{N}$, for $\tau = 1,\ldots,T$.
This frame-wise design enforces a fixed per-frame token budget, 
reducing visual tokens to a constant count for each frame and preventing anchor selection from collapsing onto a few frames.
The ACM modules comprises two sub-modules;
see its overall architecture in Fig.~\ref{fig:proposed_modules}(c).

{\bfseries Importance-based token categorization.}
Via the importance-based token categorization block in Fig.~\ref{fig:proposed_modules}(c),
at each frame $\tau$, 
we rank the visual tokens in $\widetilde{\mb{Z}}_{\tau}$ with their importance scores $\mb{s}_{\tau}$, 
and categorize the top-$K$ tokens as \dquotes{anchor} tokens $\mb{A}_\tau \in \mathbb{R}^{K \times D}$ and the remaining $N{-}K$ tokens as \dquotes{context} tokens $\mb{C}_\tau \in \mathbb{R}^{(N-K) \times D}$, where $K$ is the number of anchors.
Anchor tokens represent critical visual evidence on which a driving model relies, whereas context tokens provide complementary but less salient information.
This scheme concentrates salient visual evidence in anchors, while context tokens carry complementary details that can be merged into anchors, thereby reducing the number of visual tokens.
For example, in a crosswalk scene, tokens for the pedestrian, lane markings, and lead vehicle are treated as anchors. 
Tokens for road texture, shadows, and background patterns are context; merging them into the corresponding anchors preserves decision-critical cues while reducing the overall token count.

{\bfseries Token merging.}
To reduce visual tokens by minimizing redundancy while preserving critical information, we propose a new module that merges context tokens into their most relevant anchors.
See the overall architecture of the proposed token merging sub-module in Fig.~\ref{fig:proposed_modules}(c).
At the $\tau$th frame, 
we first compute the similarity between the anchor tokens $\mb{A}_\tau$ and context tokens $\mb{C}_\tau$. 
To this end, we embed anchor tokens as queries (Q) and context tokens as keys (K) and values (V):
\be{
\mb{Q}_\tau = \widehat{\mb{A}}_\tau \mb{W}_\text{Q}, 
~~
\mb{K}_\tau = \widehat{\mb{C}}_\tau \mb{W}_\text{K}, 
~~
\mb{V}_\tau = \widehat{\mb{C}}_\tau \mb{W}_\text{V},
~~
\forall \tau,
\label{eq:qkv_proj}
}
where 
$\widehat{\mb{A}}_\tau = \operatorname{LayerNorm} (\mb{A}_\tau)$ and $\widehat{\mb{C}}_\tau = \operatorname{LayerNorm} (\mb{C}_\tau)$,
$\mb{W}_\text{Q},\mb{W}_\text{K},\mb{W}_\text{V} \in \bbR^{D\times D_{\text{head}}}$ are learnable projections (shared across frames), 
and $D_{\text{head}}$ denotes the per-head dimension in the multi-head attention mechanism \cite{ViT}, 
$\tau = 1,\ldots,T$. 
Consequently, $\mb{Q}_\tau\!\in\!\mathbb{R}^{K\times D_{\text{head}}}$ and $\mb{K}_\tau,\mb{V}_\tau\!\in\!\mathbb{R}^{(N-K)\times D_{\text{head}}}$.

To encode anchor–context relationships as (near) hard assignments, we construct an assignment matrix 
$\mb{M}_\tau \in \mathbb{R}^{K \times (N-K)}$, where each column corresponds to a context token and its entries give the (approximately one-hot) assignment probabilities over anchors.
We adopt the Gumbel–Softmax operator, inspired by \cite{GroupVit}:
\be{
\mb{M}_\tau
=
\operatorname{Gumbel\mbox{-}Softmax}
\!\big(
D_{\text{head}}^{-1/2} \cdot \mb{Q}_\tau \mb{K}_\tau^\top 
\big), 
\quad \forall \tau,
\label{eq:gumbel}
}
where $\mb{Q}_\tau$ and 
$\mb{K}_\tau$ are the projected queries (anchors) and keys (context), respectively.
The similarity matrix $\mb{Q}_\tau \mb{K}_\tau^\top$ is scaled by $\sqrt{D_{\text{head}}}$ for numerical stability.
Unlike standard softmax -- that yields soft, distributed assignments over anchors -- Gumbel–Softmax provides a differentiable approximation to categorical sampling \cite{GumbelSoftmax}, producing columns of $\mb{M}_\tau$ that are nearly one-hot so that each context token is assigned to a single anchor.

To assign each context token to a single anchor, 
we convert each column of $\mb{M}_\tau$ into a one-hot vector by taking $\argmax$ over anchors.
Because this mapping is non-differentiable, 
we adopt the straight-through estimation (STE) approach \cite{STE}: 
we use the one-hot matrix in the forward pass, 
but in backpropagation we pass gradients as if the soft scores $\mb{M}_\tau$ had been used. 
This enables end-to-end training while enforcing (near) one-hot anchor assignments for each context token.
We denote the resulting STE-approximated hard-assignment matrix by $\widehat{\mb{M}}_\tau$.

Finally, we update the anchor representations so that the resulting tokens, $\widetilde{\mb{A}}_\tau \in \mathbb{R}^{K\times D}$, encode both their original salient signal and the complementary cues contributed by assigned context tokens.
We realize this with the following residual merge-and-project model:
\be{
\widetilde{\mb{A}}_\tau
=
\mb{A}_\tau 
+ 
( \widehat{\mb{M}}_\tau\,\mb{V}_\tau ) \mb{W}_\text{O},
\quad
\forall \tau,
\label{eq:merged_output}
}
where $\widehat{\mb{M}}_\tau$ is the STE-approximated hard-assignment matrix given above, 
$\mb{V}_\tau$ are context values given in \R{eq:qkv_proj}, and 
$\mb{W}_\text{O} \in \bbR^{D_{\text{head}}\times D}$ is a learnable projection.
The product $\widehat{\mb{M}}_\tau \mb{V}_\tau$ aggregates, for each anchor (row), all context tokens assigned to it, and the projection $\mb{W}_\text{O}$ maps the merged features back to the embedding dimension $D$.
The resulting $\widetilde{\mb{A}}_\tau$ in \R{eq:merged_output} constitutes the reduced set of visual tokens for each frame.

\subsubsection{LLM}
\label{sec:LLM}

We aggregate the reduced visual tokens in \R{eq:merged_output} from $T$ recent frames -- 
specifically,
$\widetilde{\mb{A}} = [\widetilde{\mb{A}}_1^\top, \cdots, \widetilde{\mb{A}}_T^\top]^\top \in \bbR^{TK \times D}$ 
--
and further concatenate $\widetilde{\mb{A}}$ with text tokens.
Finally, we feed the reduced visual-text tokens to a frozen LLM to produce the output tokens for predicting waypoints.

\subsection{Proposed auxiliary path for training}
\label{sec:auxi_path}

To train the importance predictor in \S\ref{sec:importance_predictor}, 
we proposes an auxiliary path that generates pseudo-supervision signals by using all visual tokens \emph{without} reduction.
We first project all visual tokens from a vision backbone, 
concatenate them with text tokens along the token dimension, 
and pass the result into the frozen LLM backbone (\S\ref{sec:LLM}).
From the last transformer decoder block in the LLM, 
we extract the attention score matrix, 
where rows correspond to query tokens and columns correspond to key tokens. 
To quantify how much attention each token receives, 
we take the column-wise mean of this matrix, resulting in a vector of pseudo importance scores. 
Each element of this vector indicates the average attention a token receives across all query positions (i.e., all text and visual tokens in the last self-attention layer), and tokens with higher pseudo-importance scores are considered more important for the downstream task.
We use the pseudo-importance scores for visual tokens as supervision signals to train the importance predictor.

\subsection{Control}
\label{sec:control}

From the output tokens of the frozen LLM (\S\ref{sec:LLM}), 
we predict future waypoints at each time step $t$.
Specifically, 
at (current) time $t$,
given past $T$ frames $t' = t - T + 1, \ldots, t$, 
the proposed model predicts $T_+$ future waypoints $\{p_{t + 1},\ldots,p_{t + T_+}\}$  via a two-layer MLP \cite{LMDrive}.
Following \cite{LMDrive}, 
we use two proportional–integral–derivative controllers to map predicted waypoints to low-level control: 
a lateral controller that tracks the trajectory heading to produce steering commands, and a longitudinal controller that regulates speed along the waypoints to produce throttle/brake commands.



\subsection{E2E training loss}
\label{sec:e2e_training}

The proposed training objective comprises two components.
First, we define the score loss $\mathcal{L}_{\text{score}}$ as the $\ell_1$ distance between pseudo-supervision signals from the auxiliary path in \S\ref{sec:auxi_path} and importance scores predicted by the 
importance predictor in \S\ref{sec:importance_predictor}.
Second, the waypoint loss $\mathcal{L}_{\text{wp}}$ is defined as the $\ell_1$ discrepancy between predicted future waypoints in \S\ref{sec:control} and ground-truth waypoints.  
The final objective is
$\mathcal{L} = \mathcal{L}_{\text{wp}} + \lambda \mathcal{L}_{\text{score}}$,
where $\lambda$ balances the two losses.
The proposed design enables end-to-end training, 
jointly learning to predict waypoints and to estimate visual token importance.

\section{Experimental results and discussion}
\label{sec:experiment}

This section describes our experimental setups and presents the results with some discussion.
\S\ref{sec:main_result} compares driving performances and inference efficiencies of three E2E driving MLLMs, each with two LLM backbones of different scales, LLaVA \cite{LLaVA} and TinyLLaVA \cite{tinyllava}:
\textit{\romnum{1})} an E2E driving MLLM using all visual tokens, hereafter referred to as \dquotes{all-token LMDrive},
\textit{\romnum{2})} SOTA E2E driving MLLM, LMDrive \cite{LMDrive},\footnote{The SOTA LMDrive setup incorporates Q-Former-based visual token reduction and LLaVA \cite{LMDrive}.}
and
\textit{\romnum{3})} proposed SToRM.
\S\ref{sec:comparison_reduction_method} compares driving performances of proposed SToRM with E2E driving MLLMs with the seven representative 
SOTA visual token reduction methods \cite{ToMe, PruMerge, DivPrune, VisionZip, BLIP2, HiRED, HICom}.
For fair comparisons, we reduced the number of visual tokens to $120$ across all the methods.
\S\ref{sec:predictor_variants}--\S\ref{sec:ACM_variants} investigate variants of proposed importance prediction in \S\ref{sec:importance_predictor} and ACM in \S\ref{sec:ACM}.

\begin{table*}[t!]
\centering
\caption{Comparisons between SToRM and SOTA E2E driving MLLM with two LLM backbones of different scales
(the symbols $^\uparrow$ and $^\downarrow$ denote that higher and lower values are better, respectively; LangAuto-Long dataset).}
\label{tab:main_result}

\vspace{-0.5pc}
\resizebox{0.95\textwidth}{!}{
\begin{tabular}{clc|ccc|ccc}
\toprule
\multirow{3}{*}{\makecell[c]{LLM \\ backbones}}
& \multirow{2}{*}{\textbf{MLLMs}}
& \multirow{2}{*} {\makecell{\textbf{\# of visual} \\ \textbf{tokens}}}
& \multicolumn{3}{c|}{\textbf{Driving performances}} 
& \multicolumn{3}{c}{\textbf{Inference efficiencies}} \\
\cmidrule(lr){4-6} \cmidrule(lr){7-9}
& & & \textbf{DS$^\uparrow$} & \textbf{RC$^\uparrow$} & \textbf{IS$^\uparrow$} 
& \textbf{TFLOPs$^\downarrow$} 
& \textbf{Memory$^\downarrow$}
& \textbf{FPS$^\uparrow$} \\
\midrule

\multirow{3}{*}{\makecell[c]{LLaVA \\ (7B)}}
 & All-token LMDrive & 3,000 
 & $44.0 \pm 2.2$ & $56.5 \pm 3.8$ & $\mb{0.82} \pm 0.02$ 
 & 30.80 & 16.20 & 4 \\

 & LMDrive & 120  
 & $36.2 \pm 2.3$ & $46.5 \pm 4.3$ & $0.81 \pm 0.03$ 
 & 1.04 & \textbf{14.10} & 24 \\

 & \cellcolor{gray!10}\textbf{SToRM (ours)}
 & \cellcolor{gray!10}120  
 & \cellcolor{gray!10}$\mb{44.2} \pm 2.5$ 
 & \cellcolor{gray!10}$\mb{56.8} \pm 3.3$ 
 & \cellcolor{gray!10}$\mb{0.82} \pm 0.02$ 
 & \cellcolor{gray!10}\textbf{1.02} 
 & \cellcolor{gray!10}\textbf{14.10}
 & \cellcolor{gray!10}\textbf{25} \\

\midrule

\multirow{3}{*}{\makecell[c]{TinyLLaVA \\ (1.5B)}}
 & All-token LMDrive & 1,800 
 & $39.6 \pm 2.1$ & $48.7 \pm 4.0$ & $0.82 \pm 0.03$ 
 & 2.65 & 3.70 & 6 \\

 & LMDrive & 120 
 & $30.9 \pm 2.6$ & $39.5 \pm 4.1$ & $0.82 \pm 0.01$ 
 & 0.18 & \textbf{3.30} & 34 \\

 & \cellcolor{gray!10}\textbf{SToRM (ours)}
 & \cellcolor{gray!10}120 
 & \cellcolor{gray!10}$\mb{40.8} \pm 1.9$ 
 & \cellcolor{gray!10}$\mb{49.3} \pm 3.5$ 
 & \cellcolor{gray!10}$\mb{0.84} \pm 0.01$ 
 & \cellcolor{gray!10}\textbf{0.16}
 & \cellcolor{gray!10}\textbf{3.30}
 & \cellcolor{gray!10}\textbf{36} \\

\bottomrule
\end{tabular}}
\vspace{-1pc}
\end{table*}

\begin{table}[t!]
\centering
\caption{Comparisons between SToRM and SOTA E2E driving MLLM (LangAuto-Short and LangAuto-Tiny datasets)}
\label{tab:langauto_short}
\vspace{-0.5pc}
\resizebox{0.98\columnwidth}{!}{\begin{tabular}{llccc ccc}
\toprule
 \multirow{2}{*}{\textbf{LLMs}} 
& \multirow{2}{*}{\textbf{MLLMs}}
&\multicolumn{3}{c}{\textbf{LangAuto-Short}} &
\multicolumn{3}{c}{\textbf{LangAuto-Tiny}} \\
\cmidrule(lr){3-5} \cmidrule(lr){6-8}
& & \textbf{DS$^\uparrow$} & \textbf{RC$^\uparrow$} & \textbf{IS$^\uparrow$} 
  & \textbf{DS$^\uparrow$} & \textbf{RC$^\uparrow$} & \textbf{IS$^\uparrow$} \\
\midrule
\multirow{2}{*}{LLaVA} 
  & LMDrive & $50.6$ & $60.0$ & $0.84$
             & $66.5$ & $77.9$ & $0.85$ \\
  & \cellcolor{gray!10}\textbf{SToRM}   & \cellcolor{gray!10}$\mb{64.5}$ & \cellcolor{gray!10}$\mb{74.7}$ & \cellcolor{gray!10}$\mb{0.88}$
             & \cellcolor{gray!10}$\mb{78.8}$ & \cellcolor{gray!10}$\mb{86.9}$ & \cellcolor{gray!10}$\mb{0.92}$ \\
\midrule
\multirow{2}{*}{TinyLLaVA} 
  & LMDrive & $43.2$ & $56.1$ & $0.83$
             & $60.5$ & $74.9$ & $0.86$ \\
  & \cellcolor{gray!10}\textbf{SToRM}  & \cellcolor{gray!10}$\mb{55.4}$ & \cellcolor{gray!10}$\mb{65.4}$ & \cellcolor{gray!10}$\mb{0.86}$
             & \cellcolor{gray!10}$\mb{75.0}$ & \cellcolor{gray!10}$\mb{82.1}$ & \cellcolor{gray!10}$\mb{0.94}$ \\
\bottomrule
\end{tabular}}
\vspace{-0.3pc}
\end{table}

\begin{table}[t!]
\centering
\caption{Comparisons b/w different token red.~methods in E2E driving}
\label{tab:token_reduction_comparison}
\vspace{-0.5pc}
\begin{tabular}{l | ccc}
\toprule
\makecell{\textbf{Reduction methods} \\ \textbf{(120 visual tokens)}} & \textbf{DS$^\uparrow$} & \textbf{RC$^\uparrow$} & \textbf{IS$^\uparrow$}  \\
\midrule
Random         & $24.3 \pm 1.8$ & $36.2 \pm 2.2$ & $0.75 \pm 0.02$ \\
ToMe \cite{ToMe}          & $28.4 \pm 1.3$ & $38.0 \pm 1.9$ & $0.75 \pm 0.01$ \\
LLaVA-PruMerge \cite{PruMerge} & $30.7 \pm 2.6$ & $39.6 \pm 2.9$ & $0.80 \pm 0.03$ \\
DivPrune \cite{DivPrune}       & $31.5 \pm 2.0$ & $39.8 \pm 2.7$ & $0.79 \pm 0.01$ \\
VisionZip \cite{VisionZip}      & $34.7 \pm 2.2$ & $44.3 \pm 2.5$ & $0.78 \pm 0.02$ \\
Q-Former \cite{BLIP2}       & $36.2 \pm 2.3$ & $46.5 \pm 4.3$ & $0.81 \pm 0.03$ \\
HiRED \cite{HiRED}         & $36.6 \pm 2.3$ & $46.7 \pm 4.0$ & $0.80 \pm 0.01$ \\
HiCom \cite{HICom}          & $37.5 \pm 2.5$ & $49.5 \pm 3.8$ & $0.78 \pm 0.02$ \\
\rowcolor{gray!10}
\textbf{SToRM (ours)}     & $\mb{44.2} \pm 2.5$ & $\mb{56.8} \pm 3.3$ & $\mb{0.82} \pm 0.02$ \\
\bottomrule
\end{tabular}
\vspace{-1pc}
\end{table}

\subsection{Experimental setups}

\textit{1) Dataset:} We used the Language-guided Autonomous Driving (LangAuto) benchmark dataset \cite{LMDrive} constructed by the CARLA simulator.
The LangAuto dataset consists of challenging driving scenarios (e.g., highways, intersections, and roundabouts) and various environmental conditions (e.g., weathers and daylight).
The LangAuto benchmark dataset includes
\textit{\romnum{1})} RGB images from front-, left-, right-, and rear-facing cameras
\textit{\romnum{2})} point clouds from a center LiDAR,
\textit{\romnum{3})} navigation instructions (e.g., lane changing, turning), and
\textit{\romnum{4})} control commands.
In constructing training, validation, test splits, we followed the setup \cite{LMDrive}.
For training, we used $51$k chunks, totaling $2.5$M frames.
For validation, we used $13$k chunks, totaling $0.6$M frames.
For testing, we used three sub-tracks with different route lengths: LangAuto-Long,  Short, and Tiny.
LangAuto-Long, -Short, and -Tiny contain routes of $>\!500\,\mathrm{m}$, $150$--$500\,\mathrm{m}$, and $<\!150\,\mathrm{m}$, respectively.
Unless otherwise specified, we report results with LangAuto-Long.

\textit{2) Evaluation metrics:}
We evaluated driving performance using route completion (RC), infraction score (IS), and driving score (DS). 
RC is the fraction of the route completed; $\text{IS} \!\in\! [0,\! 1]$ quantifies agent-triggered infractions; and DS captures both progress and safety as $\text{RC} \!\times\! \text{IS}$, reported as a percentage.
We report inference efficiency in terms of floating-point operations (FLOPs; $1\,\text{GFLOP} = 10^{9}\,\text{FLOPs}$; $1 \,\text{TFLOP} = {10}^{12}\,\text{FLOPs}$), peak memory usage in gigabytes (GB) 
and frames per second (FPS), measured on a single NVIDIA RTX 4090 GPU.
We report averages over three runs for all metrics.

\textit{3) Implementation details:}
Following \cite{LMDrive}, we used $\{ T=30, N=100, D=256, T_+=10\}$.
For sliding windows of importance predictor in \S\ref{sec:importance_predictor}, we set 
$\{ \ell=1, \kappa=2, L=4 \}$.
For ACM module in \S\ref{sec:ACM}, we set 
$\{ K=4, D_\text{head}=64 \}$.
For training, we followed the hyperparameters and optimizer chosen in \cite{LMDrive}.
For the proposed loss in \S\ref{sec:e2e_training}, we set $\lambda$ as $50$.
We ran all experiments with four NVIDIA A100 GPUs and the CARLA simulator (ver.~0.9.10).


\subsection{Comparisons between SToRM and SOTA E2E driving MLLM with different LLM scales}
\label{sec:main_result}

Across LLM backbones of different scales,
Table~\ref{tab:main_result} shows that the proposed SToRM achieves very comparable driving performances with SOTA E2E
driving MLLM \cite{LMDrive} using all tokens, i.e., all-token LMDrive, with far lower computational cost.
In particular, compared with all-token LMDrive, 
SToRM reduces FLOPs by approximately $30\times$ with the large LLM backbone and $16.6\times$ with the tiny backbone.
These reductions make real-time E2E driving inference feasible on a standard GPU, with over $25$ FPS.
Tables~\ref{tab:main_result}-\ref{tab:langauto_short} show that
under the same visual-token budget, 
SToRM outperforms SOTA model \cite{LMDrive},
indicating efficient token use and robust generalization across LLM sizes.
Finally, 
Table~\ref{tab:main_result} shows that
compared with \emph{the} SOTA model, LMDrive (using LLaVA + Q-Former token reduction) \cite{LMDrive}, 
SToRM with the tiny LLM backbone, TinyLLaVA, attains significantly better driving performance while substantially improving inference efficiency, specifically,
$6.5\times$ reduction in FLOPs, $4.27\times$ less memory, and $1.5\times$ higher FPS.

\subsection{Comparisons between different visual token reduction methods in MLLM-based E2E driving}
\label{sec:comparison_reduction_method}

Table~\ref{tab:token_reduction_comparison} demonstrates that with the E2E driving MLLM architecture held fixed, the proposed SToRM outperforms the representative SOTA visual token reduction baselines for E2E driving.
This implies that the proposed task-relevant supervision for token reduction is more effective than heuristic, e.g., similarity-based criteria, particularly in E2E driving.

\subsection{Comparisons b/w different importance predictor designs}
\label{sec:predictor_variants}

To evaluate architectural choices for importance prediction, 
this section compares our lightweight MLP-Mixer-based predictor in \S\ref{sec:importance_predictor} with a Transformer-based alternative (using four encoding blocks \cite{Transformer}), 
with and without the proposed sliding window mechanism in \R{eq:window_tokens}.
Table~\ref{tab:score_predictor_comparison} shows that 
the proposed lightweight MLP-Mixer-based predictor achieves comparable driving performance to a Transformer-based alternative while substantially reducing computational cost, 
under both sliding-window and non-sliding configurations.
Moreover, the proposed sliding-window mechanism in \R{eq:window_tokens} significantly reduces computation without degrading driving performances, implying that short-term windows capture sufficient temporal context -- obviating the need to process all visual tokens from the $T$ most recent frames.




\begin{table}[t!]
\centering
 \caption{Comparisons b/w different importance predictor designs}
\label{tab:score_predictor_comparison}
\vspace{-0.5pc}
\resizebox{\columnwidth}{!}
{\begin{tabular}{cc|ccc|c}
\toprule
\makecell{\textbf{Block} \\ \textbf{architectures}} & \makecell{\textbf{Our sliding} \\ \textbf{window mech.}} & \textbf{DS$^\uparrow$} & \textbf{RC$^\uparrow$} & \textbf{IS$^\uparrow$} & \textbf{GFLOPs$^\downarrow$} \\
\midrule
Transformer  & $\times$     & $\mb{46.1}$ & $\mb{59.8} $ & $\mb{0.82} $ & 11.3 \\
Transformer  & $\bigcirc$   & $45.0$ & $58.1 $ & $0.80 $ & 5.4 \\
MLP-Mixer  & $\times$     & $45.7$ & $59.6 $ & $0.81 $ & 6.2 \\
\rowcolor{gray!10} \textbf{MLP-Mixer}  & $\boldbigcirc[1.05ex][1pt]$   & $44.2$ & $56.8 $ & $\mb{0.82} $ & $\mb{2.1}$ \\
\bottomrule
\end{tabular}}
\vspace{-0.3pc}
\end{table}

\begin{table}[t!]
\centering
\caption{Comparisons b/w different token reduction schemes in ACM}
\label{tab:ACM_variants}
\vspace{-0.5pc}
\resizebox{0.8\columnwidth}{!}{%
\begin{tabular}{l|ccc}
\toprule
\textbf{Reduction schemes} & \textbf{DS$^\uparrow$} & \textbf{RC$^\uparrow$} & \textbf{IS$^\uparrow$} \\
\midrule
Only anchor tokens & $41.8$ & $53.6$ & $\mb{0.83}$ \\
Soft merging   & $40.1$ & $51.2$ & $0.80$ \\
\rowcolor{gray!10} \textbf{Hard merging (ours)}   & $\mb{44.2}$ & $\mb{56.8}$ & $0.82$ \\
\bottomrule
\end{tabular}}
\vspace{-1.2pc}
\end{table}

\subsection{Comparisons between different visual token reduction approaches in ACM module}
\label{sec:ACM_variants}

This section studies different token reduction schemes in the proposed ACM module in \S\ref{sec:ACM}:
\textit{\romnum{1})} only use top-$K$ anchor tokens from our importance-based token categorization without merging, 
\textit{\romnum{2})} a \dquotes{soft} alternative that merges context tokens into \emph{all} anchors with weighted contributions, and 
\textit{\romnum{3})} the proposed \dquotes{hard} scheme in \R{eq:merged_output}, which merges each context token into its single most relevant anchor.
Table~\ref{tab:ACM_variants} shows that 
the proposed hard-assignment merging scheme outperforms the top-$K$-only selection scheme and 
the soft, naïve attention-based merging alternative.
This suggests that hard merging preserves informative anchor representations, whereas soft merging introduces over-smoothing by merging contributions from all context tokens.

\section{Conclusion}
\label{sec:conclusion}

E2E autonomous driving models process multi-modal sensor inputs and can benefit from language guidance. 
However, MLLM-based designs are computationally heavy due to large LLM backbones and many multi-modal tokens.
It is critical to reduce compute for resource-constrained vehicles.

We proposed SToRM, the first supervised token reduction framework for MLLMs that can significantly reduce computational cost \emph{without} degrading E2E driving performance.
The proposed SToRM outperformed the SOTA E2E driving MLLM \cite{LMDrive} in both driving performances and inference efficiencies on the LangAuto benchmark.
The key idea of SToRM is to leverage pseudo-supervision signals to guide importance-aware token reduction. 
SToRM employs a lightweight importance predictor and an ACM module that merges less important tokens into anchor tokens for efficient processing.
\vspace{-1pc}

\bibliographystyle{IEEEtran}
\bibliography{ref}
\addtolength{\textheight}{-12cm} 
\end{document}